\documentclass[11pt]{article}
\usepackage{arabtex}
\usepackage{coling2020}
\usepackage{times}
\usepackage{url}
\usepackage{latexsym}
\usepackage[utf8]{inputenc}
\usepackage{hyperref}

\usepackage{utf8}
\usepackage{graphicx}
\usepackage{amsmath}
\usepackage[utf8]{inputenc}
\setcode{utf8}
\usepackage{multirow}
\usepackage[normalem]{ulem}
\useunder{\uline}{\ul}{}
\usepackage[titletoc,toc,title]{appendix}
\colingfinalcopy 

\title{Is it Great or Terrible? Preserving Sentiment in Neural Machine Translation of Arabic Reviews }

\author{Hadeel Saadany \\
  Centre for Translation Studies \\
  University of Surrey, UK \\
  {\tt hadil.saadany@gmail.com} \\\And
  Constantin Or\u{a}san \\
  Centre for Translation Studies \\
  University of Surrey, UK \\
  {\tt C.Orasan@surrey.ac.uk} \\}

\date{}

\begin{document}
\maketitle
\begin{abstract}
   Since the advent of Neural Machine Translation (NMT) approaches there has been a tremendous improvement in the quality of automatic translation. However, NMT output still lacks accuracy in some low-resource languages and sometimes makes major errors that need extensive post-editing. This is particularly noticeable with texts that do not follow common lexico-grammatical standards, such as user generated content (UGC). In this paper we investigate the challenges involved in translating book reviews from Arabic into English, with particular focus on the errors that lead to incorrect translation of sentiment polarity. 
   Our study points to the special characteristics of Arabic UGC, examines the sentiment transfer errors made by Google Translate of Arabic UGC to English, analyzes why the problem occurs, and proposes an error typology specific of the translation of Arabic UGC. Our analysis shows that the output of online translation tools of Arabic UGC can either fail to transfer the sentiment at all by producing a neutral target text, or completely flips the sentiment polarity of the target word or phrase and hence delivers a wrong affect message. 
   We address this problem by fine-tuning an NMT model with respect to sentiment polarity showing that this approach can significantly help with correcting sentiment errors detected in the online translation of Arabic UGC.
  
\end{abstract}

\section{Introduction}
\label{intro}

%
%
\blfootnote{
    %
    %
    %
    %
    \hspace{-0.65cm}  
    This work is licensed under a Creative Commons 
    Attribution 4.0 International Licence.
    Licence details:
    \url{http://creativecommons.org/licenses/by/4.0/}.
    %
    %
}

Translation of user generated content (UGC) such as user reviews is becoming common on multilingual websites which sell products and services such as amazon.com or booking.com. In this context, sentiment preservation in automatic machine translation (these days usually neural machine translation (NMT) output) is of great importance because many decisions about purchasing a product or service are based on the comments made by others. 
There have been different studies which explored the transfer of sentiment in MT, but most of these studies assess how far automatic sentiment classification systems can capture sentiment information from the translations \cite{trans_for_low,evaluation,impact}. The objective of most research in this area is from a sentiment classification perspective rather than a translation accuracy perspective. Hence, it measures how far automatic translation of a language into English can help with the sentiment classification of that language by applying the available English sentiment resources on the target text \cite{crosslingual,arabic_senti_analysis,howtrans,withoutgood}. 

This study is concerned with NMT accuracy of sentiment transfer at the word/phrase level and shows that inaccurate translation can transfer a completely opposite affect message. Moreover, the translation of UGC such as product reviews constitutes a significant challenge for NMT online tools in general and for Arabic UGC in particular. The reason is that Arabic UGC is usually a mix of Dialectical Arabic (DA) and Modern Standard Arabic (MSA) which differ significantly on the lexico-grammatical level. The same word or phrase can have opposite sentiment polarities in the two versions of the Arabic language, which often leads to a mistranslation of the sentiment message. If the NMT engine is robust enough to handle this type of code-switching, it can become more reliable not only in downstream NLP tasks such as cross-lingual information retrieval, but also in real-life scenarios when Internet users resort to online translation tools to check the reviews of a particular product of interest. In this study, we assess the degree to which the NMT online tools transfer sentiment accurately at the word/phrase level and suggest methods for improving the accuracy of the translation of sentiment in Arabic UGC. We aim to answer the following questions:

\begin{enumerate}
\item What type of errors in the output of NMT of Arabic UGC cause problems in sentiment preservation?
\item How can a sentiment sensitive input for an NMT model help with a more accurate sentiment polarity transfer of Arabic UGC?
\item 	How can sentiment preservation in the target language be measured and whether the BLEU score is the most appropriate metric for  evaluating translation of sentiment?

\end{enumerate}

To answer the above research questions, this paper is divided as follows: section 2 presents related work on sentiment transfer in MT. Section 3 analyzes sentiment translation errors of NMT online tools of Arabic reviews and provides a qualitative typology of most frequent error types. In section 4, we present different methodological approaches for correcting the NMT online sentiment transfer errors. Section 5 provides task-specific evaluation metrics for assessing the sentiment accuracy improvement by the proposed methods. Section 6 presents a conclusion on the different experiments as well as limitations of the present study.

\section{Related Work}

Research on the translation of sentiment in MT has focused on the idea that despite significant errors in sentiment transfer, automatic sentiment classification systems are still able to capture sentiment information from the translations \cite{crosslingual,impact,howtrans,arabic_senti_analysis}. Salameh et al. \shortcite{senti_after} showed that although certain attributes of automatically translated text `may mislead humans' with regards to the true sentiment of the source text, they do not seem to affect the automatic sentiment analysis systems \cite{senti_after}. The rationale behind these studies is that if we have a good machine translation model, it will  eliminate the necessity to develop sentiment analysis resources specific of the source language \cite{trans_for_low}. Given the proliferation of English sentiment analysis tools, we can always make use of them by conducting sentiment analysis on the English translation of the source text, even if the translation is not of high quality \cite{withoutgood,evaluation}. Studies also show that developed MT models, as well as online translation tools such as Google Translate and Microsoft Translate, can be relied upon to perform sentiment classification of the target text despite any accuracy errors \cite{impact}.  This is because sentiment classification systems can learn an appropriate model even from mistranslated text — especially when automatic translation makes consistent errors \cite{senti_after}. Moreover, statistically, studies of sentiment translation have shown that automatic translation leads to only about 60\% match with manually annotated sentiment labels. Yet, automatic sentiment classifiers can still perform well despite these errors which can markedly impact human perception of sentiment in the source tweet/review \cite{senti_after,howtrans}.


Recently, MT studies started to tackle how sentiment can be preserved in the translation of UGC from a translation accuracy perspective. B\'erard et al. \shortcite{restaurant} show that back translation of restaurant reviews can provide significant improvement over existing online systems particularly in preserving sentiment of  translated UGC. They translate a large corpus of reviews from the target language into English and then use it in model training. They use domain tags at the training stage to distinguish user-generated source text. Their results prove that both synthetic data and domain tags can achieve good results in preserving the affect polarity on the sentence level. While their model is promising, they still point to serious errors in the translation of UGC such as missing negations, hallucinations, unrecognized named entities and insensitivity to context. They suggest that this task is far from solved \cite{restaurant}.

Lohar et al. \shortcite{balancing} makes an attempt to improve the sentiment transfer of translated tweets. They show that freely available translation tools often cause the sentiment encoded in the original tweet to be altered. As a consequence, they build separate negative, neutral and positive sentiment SMT models to improve sentiment preservation in the target language. They show that a translation model specific of each sentiment pole provides much better results over a single baseline model trained on the whole twitter data, regardless of the sentiment class. They attempt to strike a balance between improving sentiment transfer and preserving translation accuracy as measured by evaluative metrics such as BLEU and METEOR \cite{balancing}.
A similar technique is used by Si et al. \shortcite{sentiaware} as they build a valence sensitive NMT model for the translation of ambiguous words that can have different polarities in different contexts. Each input sentence is annotated with a positive or negative label to indicate its polarity. They show that adding this tag to the source sentence at the training time and creating dual polarity embedding vectors for ambiguous words can improve sentiment transfer at the word level \cite{sentiaware}. 

There has also been some research on finding alternative means for assessing the transfer of sentiment in MT other than the typical accuracy metrics.  B\'erard et al. \shortcite{restaurant} show that automatic evaluation metrics such as BLEU and METEOR tend to neglect sentiment discrepancies between source and target output. They suggest assessing the accuracy of sentiment preservation by targeted metrics that measure how well polysemous words are translated, or how well sentiments expressed in the original text can be recovered from its translation \cite{restaurant}. To assess sentiment preservation in MT, Lohar et al. \shortcite{maintaining} use a sentiment lexicon-based measure in combination with regular evaluation metrics such as the BLEU score. Several studies also resort to human evaluation to measure how far a model improves sentiment transfer at the word/phrase level \cite{sentiaware,howtrans}.

In this study, we evaluate the preservation of sentiment in translation not as a sentiment classification task, but from a translation accuracy perspective. We show that translation
inaccuracies at the word/phrase level can seriously impact the transfer of sentiment in Arabic UGC, which can lead to problems for users of the MT tools in real-life situations. Several commercial global platforms rely on publicly available MT engines to translate product reviews into the customers own language to facilitate communication between partners and customers\footnote{{For example, \href{https://www.booking.com/}{Booking.com}} uses Google API to translate reviews on hotels for customers on the fly.}. Inaccurate translation of reviewers' sentiment would defeat the purpose of using such tools. Moreover, in commercial situations, companies may want to find out what their users think of particular products so the accuracy of each translation review counts. Broadly speaking, online tools such as Google Translate, are commonly utilized as an off-the-shelf solution for the translation of UGC in Arabic as well as in other languages.  Error-analysis of sentiment translation by online tools, however, has proved that the true sentiment of Arabic reviews can be either missed or flipped to its exact opposite pole.

\section{Error Analysis}

In order to measure how accurately NMT online tools transfer sentiment of Arabic UGC, we chose a dataset of book reviews scraped from Goodreads\footnote{\url{https://www.goodreads.com}}   \cite{aly2013labr}. Each review has a rating between 1-5 assigned by its author. The language of the reviews is a mix of MSA and DA, with the largest majority of DA reviews in the Egyptian dialect. Reviews in the dataset are of varying lengths, but a large number of them have over than 100 tokens. Long reviews were split to a maximum of 20 tokens per review. After splitting, the data amounted to  about 230,000 sentences. This dataset was translated into English using the Google Translate API and was analysed using both manual and automatic error analysis, focusing on mistranslation of sentiment. 
Automatic sentiment analysis tools were utilized to detect sentiment errors in the dataset and subsequently select a sample for manual error analysis.

Since the main objective is to assess the accuracy of sentiment translation at the word level, we used an automated lexicon-based sentiment measure on the Google Translate output. We applied the cloud-based Microsoft Azure Text Analytics tools for sentiment analysis \footnote{\href{https://azure.microsoft.com/en-gb/services/cognitive-services/text-analytics/}{Microsoft Azure Text Analytics}} on around 13,000 target sentences. The Azure’s Sentiment Analysis API  generates sentiment scores using classification features such as \mbox{n-gram} sentiment scores, part-of-speech tags and word embeddings. It evaluates text and returns a label (positive, neutral, negative) for each sentence as well as numeric confidence scores that range from \mbox{0 to 1} for each sentiment category. Scores closer to 1 indicate a higher confidence in the label's classification, while lower scores indicate lower confidence. For each sentence, the predicted scores associated with the labels (positive, negative and neutral) add up to 1. Following traditional methods in sentiment classification \cite{thumbs}, we used the rating of the book review as indicative of its sentiment polarity and compared it to the confidence scores generated by the Azure Sentiment Analysis API. Accordingly, reviews were categorized based on discrepancies between the ratings and the confidence scores. A positive review that had a rating of 4 or above and an English negative sentiment score of 0.5 and above was extracted as an example of potential wrong negative polarity in the target text. Similarly, reviews with negative ratings of 2 and below and a positive English sentiment score of 0.5 and above were extracted as instances of potential wrong positive polarity in the target text. This amounted to a total of around 4,000 potentially negative sentiment errors and around 2,000 of potentially positive sentiment errors.

A sample of reviews of 1000  parallel sentences from the dataset that had discrepancies between the automatic sentiment score and the review rating were manually analyzed to detect reasons for these discrepancies. By analyzing the causes of mistranslation of sentiment in this sample, the mistakes were categorized into a five group typology. The typology of sentiment translation errors are summarized in the following sections. One or two examples for each type of errors will be mentioned  in the following sections. The table in appendix \ref{arrtable} gives more examples of each type.

\subsection{Contronyms}

Manual analysis of the data revealed that the first type of errors which distorts the reviewer's affect message is mistranslations of  contronyms. These are words used both in DA and MSA which can have the exact opposite sentiment polarity in each of the two language varieties or in the same variety but in different contexts. For example, the word `\<رهيبه >' means `terrible' in MSA, but in DA it often means `great'. This word was frequently mistranslated as `terrible’ in the reviews dataset, causing a distortion of the sentiment of the source text. For example, the review  ` \<الروايه رهيبه عيبها الوحيد الجزء الاخير   > ' is translated as ``The narration is terrible, its only flaw is the last part". The correct translation, however, is `The novel is great, its only flaw is the last part'. Even when the infrequent positive use of the contronym is used in Arabic MSA context it is flipped to a negative pole in the translation. For example, in the review `\<ثم قال هذه الكلمه الرهيبه اقرأ >' (then he said this magnificent word: Read) the word `\<رهيبه  >' is used positively to mean `great' or  `magnificent'. The automatic translation, however, flips it to the more common negative sense by translating the review as `then he said this terrible word: Read'. Similarly, the negation of these contronyms is often mistranslated and hence alters the sentiment message of the source text. For example, the low-rated review `\< ادب الكاتب مش الفظيع  >' (the writer's literature is not that great) has the negated contronym `\< الفظيع> ' which can either mean `not terrible' or `not great' in MSA and DA respectively. The review was mistranslated as `the writer's literature is not terrible' which had a positive sentiment score whereas the original review had a low rating.

Another example is the word `\< جامد >'. In DA, it means `great' or `awesome,' whilst in MSA it refers to its literal meaning, i.e. `rigid'. Reviews stating ` \< كتاب جامد جدا> ’ (a very good book) were constantly mistranslated as `a very rigid book’ which incorrectly reflected a negative sentiment score. A list of contronyms that caused sentiment inconsistencies between source and target text was identified by the manual analysis of the sample dataset and extracted from the larger dataset of the Goodreads reviews (see appendix \ref{arrtable} for more examples of this type of error).

\subsection{Diacritic Errors}

The vowels in the Arabic language are realized by diacritics which indicate the pronunciation of the word. The same word can have different meanings based on the diacritic marks assigned, since a change in a diacritic is a change of a vowel sound. Arabic UGC is usually lacking diacritics since Arabic native speakers can easily guess which diacritic mark is intended based on the context of the word. Automatic translation, however, often fails to realize the different meanings of words if diacritics are missing and this can lead to a wrong sentiment polarity. For example, `\< من اظرف ما قرات>’ (One of the nicest things I've read) is translated as ‘The envelope of what I read’. This is because the word `\< اظرف>’ can either mean `the nicest’ or `most entertaining' if it has a `fatha’ (a short /a/) on the third letter or `envelopes’ if it has `Damma’ (a short /u/ as in ``you’’) on the same letter. Moreover, absence of diacritics causes a confusion between the transitive and intransitive use of sentiment adjectives. For example, the adjective `\< متعبه >'    can either mean `tired' if the diacritic `fatha' (a short /a/) is on the third letter or it can mean `tiring' if the diacritic `kasrah' (short /i/) is on the same letter. Thus, for example, a book review with a positive rating starting with `\< متعبه هذه الروايه>' is mistranslated as `Tired of this narration'. The correct translation of the adjective is `This novel is tiring' where the reviewer is referring to the intellectual depth of the novel. Diacritic errors as such  cause a misinterpretation of reviewer's sentiment stance. More examples are given in Appendix \ref{arrtable}.

\subsection{Idiomatic Expressions}
Idiomatic expressions both in MSA and DA are consistently mistranslated in the dataset which leads to a complete miss of the sentiment message in the review. For example, the MSA phrase `\<خفيف الظل  >' is an idiom used to describe a `funny' animate or inanimate noun. The idiom in the positive review `\< كتاب خفيف الظل  > ' (a funny book) is mistranslated as `a light-shaded book'. The target text incorrectly reflects a neutral sentiment rather than the correct positive one. This idiom's counterpart in DA `\<دمه خفيف  >' (funny) is also constantly mistranslated in the dataset. The review `\<كتاب دمه خفيف جدا  >' (
the book is very funny) is mistranslated as `his blood book is very light'. The manual analysis of the dataset set showed that, generally speaking, idioms, either in MSA or DA, constituted a challenge to the online automatic translation tool. A large number of idioms were literally translated which did not only affect the sentiment preservation of the source text, but often produced nonsensical target text.
For example, the MSA phrase `\< وهل يخفى القمر>’ is an idiom used to describe something that is unquestionably commended by the speaker. If the idiom is used in reference to a book, a good human translation would be: `It really shines through’. The Google Translate gives a literal translation -- ` Is the moon hidden?’-- which flips the sentiment polarity of the review from highly positive to neutral. (See Appendix \ref{arrtable} for more examples).

\subsection{Dialectical Expressions}

Research studies have shown that dialectical Arabic presents several challenges to MT in general \cite{mtforarabicdialects}. It was also observed from the manual analysis of the sample data that dialectical expressions constituted a special challenge for the preservation of sentiment in the source text. Arabic UGC is acceptably written in DA or MSA or a mix of both in the same text. A large number of DA sentiment expressions were either completely missed in the translation or mistranslated. For example, positive adjectives such as `\<هايل >' (great), or negative adjectives such as  `\< عبيط  >' (silly) were mostly mistaken for proper nouns and transliterated into non-English words (Hayel, Abit). In some instances, the translation was a complete opposite of the intended affect message (e.g. `\< من الجمل العبيطه المنتشره  >' (one of the widespread silly sentences) was translated as `one of the popular sentences spread ' (see more example in Appendix \ref{app:foobar}).

\subsection{Negation}

Another type of sentiment errors that is also associated with the use of DA in Arabic UGC is the mistranslation of DA negation markers. Different Arabic dialects often treat negative particles as clitics, and hence a letter is added to the stem of the word to change it to negative \cite{emad}. The majority of DA in the dataset belongs to the Egyptian dialect where negation is realized by the morpheme  `\<مـِش >' (mish) which is either placed in front of the verb or preposition, or wrapped around it \cite{negation}. From the analysis, it was found out that the translation frequently either misses the negation and hence flips the phrase to the opposite sentiment pole or mistranslates the negated phrase all together. For example, in the review ‘\< معجبنيش ان بطل الروايه ضعيف الشخصيه  >’  (I didn’t like that the protagonist of the novel has a weak character) the negation  is missed and hence the online translation output is `I admire that the protagonist of the novel is weak in character'. There are several similar instances where the mistranslations of the DA negative structure switches the sentiment to its opposite pole (see more examples in appendix \ref{arrtable}).

\section{Notes on Error Typology}

\begin{figure}[t]
\centering
\includegraphics[scale=0.40,trim={0.3cm .2cm .3cm .1cm},clip]{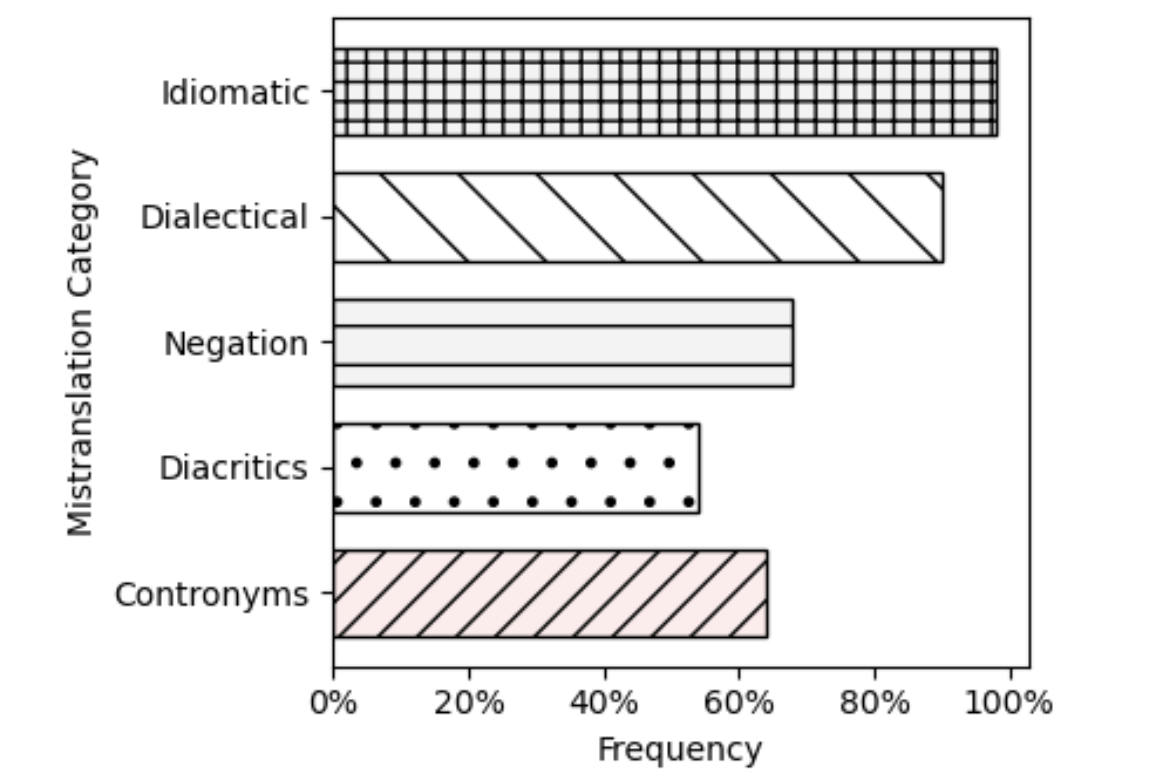}
\caption{Frequency of Error Types}
\label{acctable}
\end{figure}

In order to get an indication of the frequency of each type of errors in the whole dataset, words/phrases belonging to each group were extracted based on their frequency in the corpus then the frequency of their mistranslation in the dataset was manually calculated. Figure \ref{acctable} shows the frequency of the mistranslation instances in the total reviews dataset of the extracted words as representative of each type of errors. As can be seen almost all the instances of the most frequent idiomatic and dialectical expressions are mistranslated.  Moreover, around 65\% of the time the positive meaning of contronyms was flipped to negative. Consequently, there were sentimentally incongruous terms  where a positive noun was described with a highly improbable negative adjective (e.g. horrible achievement, terrible masterpiece, terrible happiness, and so on). 

Our error typology showed that more than one type of errors is due to code-switching between DA and MSA in Arabic UGC. There have been several approaches to tackle the challenges of translating Arabic DA such as paraphrasing source text in MSA before translation \cite{pivoting1}. Other studies have also proved that concatenating small amount of Arabic dialectical data can significantly improve the translation quality \cite{mtforarabicdialects}. However, our error analysis has shown that even in MSA context the sentiment of words/phrases can be mistranslated. Pivoting on MSA can solve straightforward problems such as dialectical phrases and dialectical idioms. However, preserving the sentiment of the source text would require addressing the polarity of words with opposite meanings such as contronyms. Since it is beyond the scope of this paper to tackle all the error types, an attempt was made to address the problem of contronyms in Arabic UGC. In this paper we propose a sentiment-sensitive NMT model that is robust to the opposite sentiment polarities of Arabic contronyms either due to code-switching between DA and MSA or to contextual variations in Arabic UGC. Details of the experiment are explained in the following sections.

\section{Sentiment Oriented NMT System}
In order to improve the translation of contronyms in Arabic UGC, we propose two transformer (NMT) models infused with sentiment information at the encoding stage. We show that training on a sentiment oriented small-sized data can provide high performance results in preserving the sentiment of challenging contronyms in Arabic UGC. Details of data preprocessing and model architectures are explained in the following sections.  

\subsection{Parallel Data Preparation and Preprocessing}


 It is worth mentioning here that the available authentic parallel English/Arabic data is mostly English to Arabic data (e.g. UN parallel corpora, TEDx scripts, and Tatoeba project \cite{arabicdata,tatoeba}). The greatest part of this data is in Arabic MSA and is not sentiment-oriented. Authentic Arabic(DA)-English parallel data in general and authentic Arabic(UGC)-English parallel data in particular is very scarce. Recently, the use of synthetic corpora in NMT led to promising results especially when authentic parallel data is scarce \cite{chinea2017adapting,cheng2020ar}. Moreover, infusing contextual cues in the input layer has proved successful in improving the robustness of the NMT models for different translation tasks even with relatively small-sized datasets \cite{johnson2017google,howsentiment,sentiaware}. Accordingly, in order to identify the correct sentiment polarity of contronyms in Arabic UGC, we opted for using the synthetic parallel data of the Goodreads reviews dataset ($\approx230,000$ sentences) for model training but with three main modifications. First, all the  mistranslation instances of the chosen list of contronyms were manually post-edited (see appendix \ref{conttable} for a list of most frequent contronyms used in the dataset). Second, the Arabic script underwent a number of preprocessing operations such as the normalization of orthographic letter forms, deletion of elongation and extra spaces. This has significantly reduced the number of out-of-vocabulary words. Third, we manually tagged all the contronyms in the source text with the right sentiment polarity according to its context. We experimented with both the tagged and the untagged post-edited source text. Details of the model architectures are in the following section. 

\subsection{NMT System Setup}

In order to explore how we can improve the translation quality, we constructed three NMT models. The first is a baseline model that takes an untagged post-edited source text as input. The baseline is a seq2seq model with an LSTM of 200 hidden states for the encoder and decoder models trained with global attention. The other two models are sentiment sensitive models that take a tagged source text as input. For the two sentiment sensitive models, we mimicked the Google Translate setup \cite{vaswani2017attention} by using a transformer for both the encoding and decoding layers with 8 heads of self-attention and with an inner feed-forward layer of size 2048, but reduced the number of training steps from 200k to 100k. We used the Adam optimizer with $\beta1 = 0.9$, $\beta2 = 0.98$ and $\epsilon= 10^9$ and the Google set up special learning rate as described by Vaswani et al. \shortcite{vaswani2017attention}. The first of the two sentiment sensitive models was initialized with random input vectors. For the second model, we created a vector space model (VSM) of the tagged source dataset where each contronym was given two distinct vectors according to its tagged sentiment polarity. A bag of words Word2Vec model was used to create the pretrained vectors of the source text \cite{gensim}. It was trained with a hierarchical softmax and a window size of 5 tokens. The pretrained word embeddings were used to initialize the second transformer model with the same parameters used for the first. All experiments were run using OpenNMT \cite{opennmt}.

\subsection{Evaluation results}

The evaluation of the proposed models was conducted on two test sets. The first was a held-out set from the Goodreads reviews ($\approx47,000$ parallel sentences). The second was a hand-crafted test set of 140 sentences where we used the list of extracted contronyms with their positive and negative sentiment connotations in an equal number of sentences and code-switched between Arabic (MSA) and Arabic (DA) either in the same sentence or among different sentences. A reference translation was created by manually translating the hand-crafted set by a native speaker. In order to adequately evaluate the performance of the models in preserving the polarity of contronyms in the source text, we conducted two types of sentiment evaluations, at the word level and at the sentence level on the held-out test set and the hand-crafted test set respectively. We compared the quality measures on both the sentence and word levels of the proposed models with the Google Translate output for the test set and the hand-crafted test set. The BLEU score was also used as a metric to assess how far the quality of the translation is balanced with the preservation of sentiment by our proposed models. Details of the experiment evaluations are explained in the next sections.

\subsubsection{Quality level}

The first evaluation metric conducted on the two datasets was based on the BLEU score. We used the metric-internal multi-detokenized BLEU \cite{bleu}. The BLEU score was used to check that the quality of the translation is not distorted while fine-tuning the models for sentiment preservation. Results in table \ref{table:1} show that both transformer models with tagged source text outperform the baseline on the two datasets. The first transformer model with tagged input, but without pretrained vectors, achieves the highest BLEU score on the test set and the negative hand-crafted test set with scores 38.77 and 44.83 respectively. The second transformer model trained on tagged source and pre-trained sentiment-oriented vectors achieves the best BLEU score on the positive hand-crafted test set (38.77). Results indicate that the overall translation quality as measured by BLEU has not been impaired with the sentiment-preservation approaches of the proposed transformer models.

\renewcommand{\arraystretch}{2}
\begin{table}[t]
\resizebox{\textwidth}{!}{%
\begin{tabular}{l|l|l|l|l|l|l|l|l|}
\cline{2-9}
 & \multicolumn{2}{c|}{\textbf{Sentence Level}} & \multicolumn{3}{c|}{\textbf{Word Level}} & \multicolumn{3}{c|}{\textbf{BLEU}} \\ \cline{2-9} 
\textbf{} & \multicolumn{2}{c|}{\textbf{Hand-Crafted   Set}} & \multicolumn{3}{c|}{\textbf{Test Set}} & \textbf{Test Set} & \multicolumn{2}{c|}{\textbf{Hand-Crafted Set}} \\ \hline
\multicolumn{1}{|l|}{\textbf{Model}} & \textbf{Positive} & \textbf{Negative} & \textbf{Precision} & \textbf{Recall} & \textbf{F1} & \textbf{} & \textbf{Positive} & \textbf{Negative} \\ \hline
\multicolumn{1}{|l|}{\textbf{Seq2seq   (no tagging)}} & .24 & .44 & 0.60 & 0.52 & 0.55 & 33.9 & 31.48 & 36.94 \\ \hline
\multicolumn{1}{|l|}{\textbf{Transformer   1 (tagging)}} & .14 & .21 & 0.74 & 0.65 & 0.69 & \textbf{38.77} & 37.56 & \textbf{44.83} \\ \hline
\multicolumn{1}{|l|}{\textbf{Transformer   2 (tagging and pre-trained)}} & \textbf{.06} & \textbf{.14} & \textbf{0.85} & \textbf{0.79} & \textbf{0.81} & 37.14 & \textbf{38.82} & 42.06 \\ \hline
\multicolumn{1}{|l|}{\textbf{Google Translate}} & .71 & .15 & 0.80 & .06 & .12  &  &  &  \\ \hline
\end{tabular}%
}
\caption{Results of Three Evaluation Metrics for Assessing Sentiment Preservation in Translation }
\label{table:1}
\end{table}
\renewcommand{\arraystretch}{1}

\subsubsection{Word-level Sentiment Evaluation}

The BLEU score can reflect the translation quality of the NMT output, but for the present study it would not be appropriate to capture how the opposite sentiments of contronyms are correctly translated. This is because the BLEU score does not give a penalty to a mistranslated sentiment lexicon that is adequately proportional to the distortion of the sentiment message. The translation of the right sentiment polarity of a contronym can be pivotal in transferring the affect message of the source text. For example, the positive use of the contronym `\<رهيبه >' in the low-rated book review `\<ليست تحفه ابداعيه رهيبه   >' (not a great creative masterpiece) is mistranslated as `not a terrible creative masterpiece' by the baseline model. The mistranslation of the contronym completely distorts the sentiment message of the review, however, the BLEU score for this mistranslation is around 76.

Accordingly, we measured the precision, recall and F1 score of the different models to assess their ability to correctly predict the true positive and true negative polarity of the contronyms in the test dataset. Table \ref{table:1} shows that the baseline model was not able to detect the correct sentiment orientation of a contronym with high accuracy, as compared to the two transformer models, despite the post-editing of the training dataset. Feeding in correct instances was not sufficient to improve the sentiment preservation of Arabic contronyms. Infusing linguistic information at the training stage, however, improved sentiment accuracy.  Moreover, the low F1 score of the  Google Translate (.12) was due to the fact that it was able to translate correctly instances of contronyms when used with negative sentiment, but failed to translate those where their positive meaning is used. Such positive cases constituted around 40\% of the instances of contronyms in the dataset. This is because the negative meaning of contronyms is more frequent in Arabic MSA, whereas the positive is used more in Arabic DA context. As explained by the error typology, Google Translate performs far better with MSA than DA.  On the other hand, the second transformer model which is trained on sentiment-sensitive pretrained vectors and tagged source text achieved best performance in depicting the true sentiment at the word level with an F1 score of .81 and a precision score of .85. The sentiment-sensitive pretrained vectors of contronyms and their polarity tagging with the second transformer model significantly helped in translating the correct sentiment at the word-level.


\subsubsection{Sentence-level Sentiment Evaluation}
The second metric for evaluating the translation of sentiment in Arabic UGC was carried out on the hand-crafted test set. In order to assess how the correct or incorrect translation of contronyms affects the total sentiment message of the source sentence, we propose a sentiment-score based metric. We compute the distance between the sentiment score of the reference sentence and the model output to measure not only how far the model preserves the sentiment of a contronym, but also the effect of translation on the sentiment context. We use the sentiment scoring methods used for error analysis, i.e. Microsoft Azure Sentiment Analysis scoring. We measure a translation cost as the mean square distance to the reference score: 
\begin{align}
    \mu_C = \frac{1}{N} \sum_{i=1}^N (s_t-s_r)^2
\end{align}
where $s_t$ is the score of the target sentence, $s_r$ is the score of the reference translation, and $N$ is the number of sentences.

As seen from table \ref{table:1} the second transformer model trained on tagged contronyms and pretrained word vectors performed best (i.e. with the lowest cost) for both the positive and negative reviews (.06, .14 respectively). It was not only more sensitive to different polarities of contronyms due to the code-switching between MSA and DA, but produced the lowest sentiment discrepancy with the sentiment scores of the reference sentence. It is also worth noting that Google Translate performed much better with the negative sense of contronyms than the positive sense. This is in line with the findings presented in the previous section. With negative contronyms, Google Translate and the second transformer model had the lowest costs of .15 and .14, respectively. However, Google Translate produced the highest discrepancy with the positive instances  (.71). Moreover, it was observed that the cost score was highest with short sentences. For instance, the positive sense of the contronym `\< رهيبه >' (awesome, great) in the short reference sentence `\< رهيبه بكل المقاييس>' (By all means awesome) is translated by Google Translate as `Terrible by all accounts'. In such cases, the sentiment cost was maximum. It is evident that if a similar distortion of sentiment messages occurs in real-life situations, it would have adverse effects on the reviewers judgement.  Examples of the output of the second transformer model (Trans2) as compared to the reference translation (Ref) and Google Translate is given in Appendix \ref{app:foobar2}.


\section{Conclusion}

This study has shown that Arabic UGC has peculiar qualities which constitute a challenge to automatic translation tools especially in its ability to preserve the sentiment message. An error typology was derived after analysing the data. This typology has highlighted how sentiment errors can impair the translation of sentiment-oriented Arabic UGC such as product reviews. Since automatic online translation tools are heavily relied upon by users and commercial platforms to translate reviews, it is of essential importance to fine-tune NMT models to the correct sentiment message in the source text. Moreover, it has been common practice for NMT training to use big parallel data which involves very high computational power and requires availability of large authentic data. The proposed NMT models in this study, however, showed that infusing contextual cues at the training stage of a relatively small data can improve the translation of sentiment in Arabic UGC both on the word and sentence level. This approach can help in providing greener training and make it feasible to construct competitive NMT tools for low-resource domains such as Arabic UGC. Moreover, we showed that translation quality metrics of sentiment-oriented Arabic UGC needs to be supplemented with other metrics that assess the preservation of sentiment. We proposed lexicon-based metrics that take into account the sentiment score of single words as well as their context. Finally, this study has tackled one of several challenges in the translation of sentiment in Arabic UGC. Future research will address the whole spectrum of challenges to improve the accuracy of sentiment preservation which is of vital importance in the translation of Arabic UGC.







\bibliographystyle{coling}
\bibliography{coling2020}

\newpage
\begin{appendices}
\section{Examples of the Error Typology for Sentiment Translation of Arabic UGC} \label{app:foobar}
\begin{figure}[h]

\centering
\includegraphics[scale=.54,trim={0cm .1cm 0cm 0cm},clip]{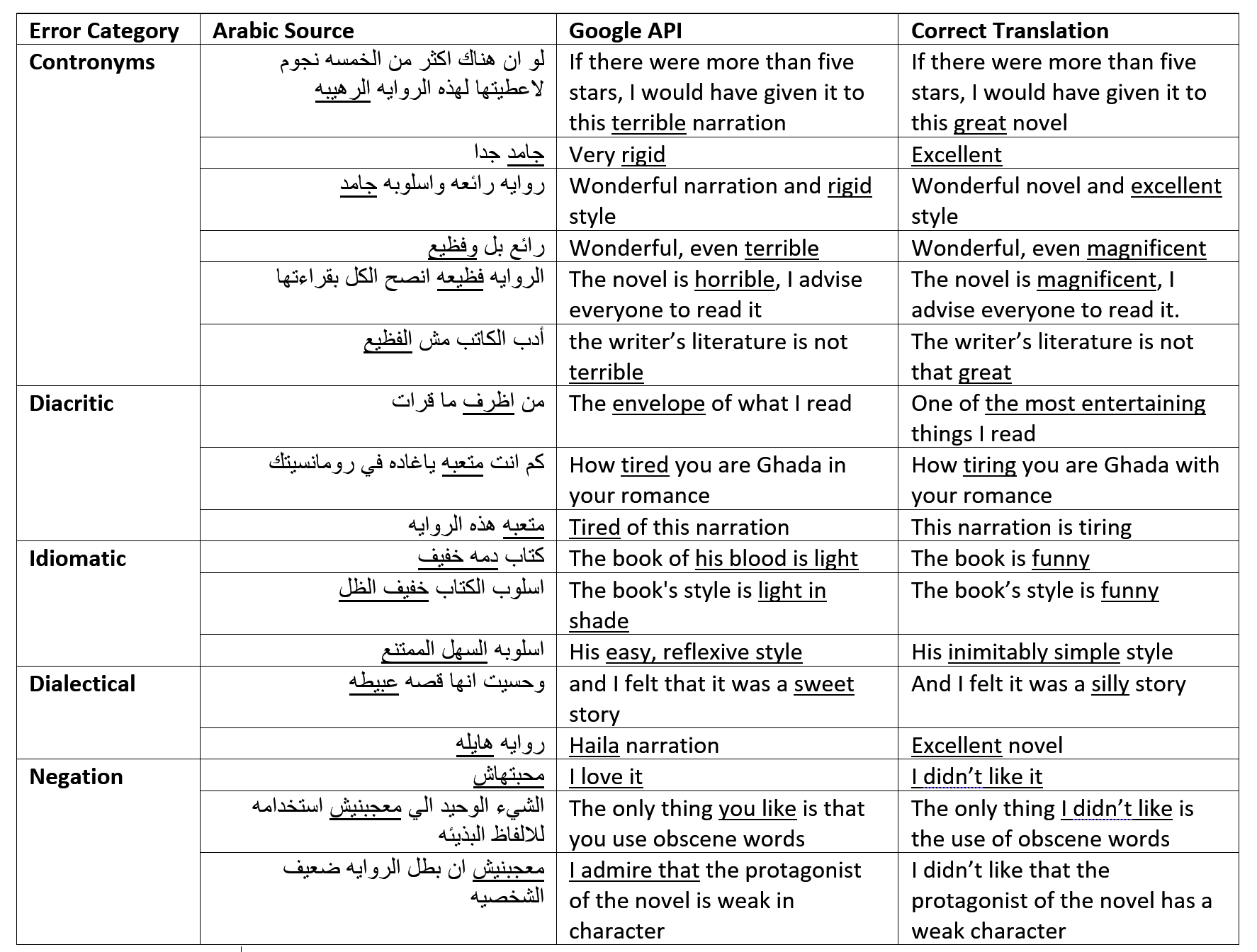}
\label{arrtable}
\end{figure}

\newpage
\section{Examples of Translation Models Output}
\label{app:foobar2}
\begin{figure}[h]
\centering
\includegraphics[scale=.55,trim={0cm .222cm 0cm .1cm},clip]{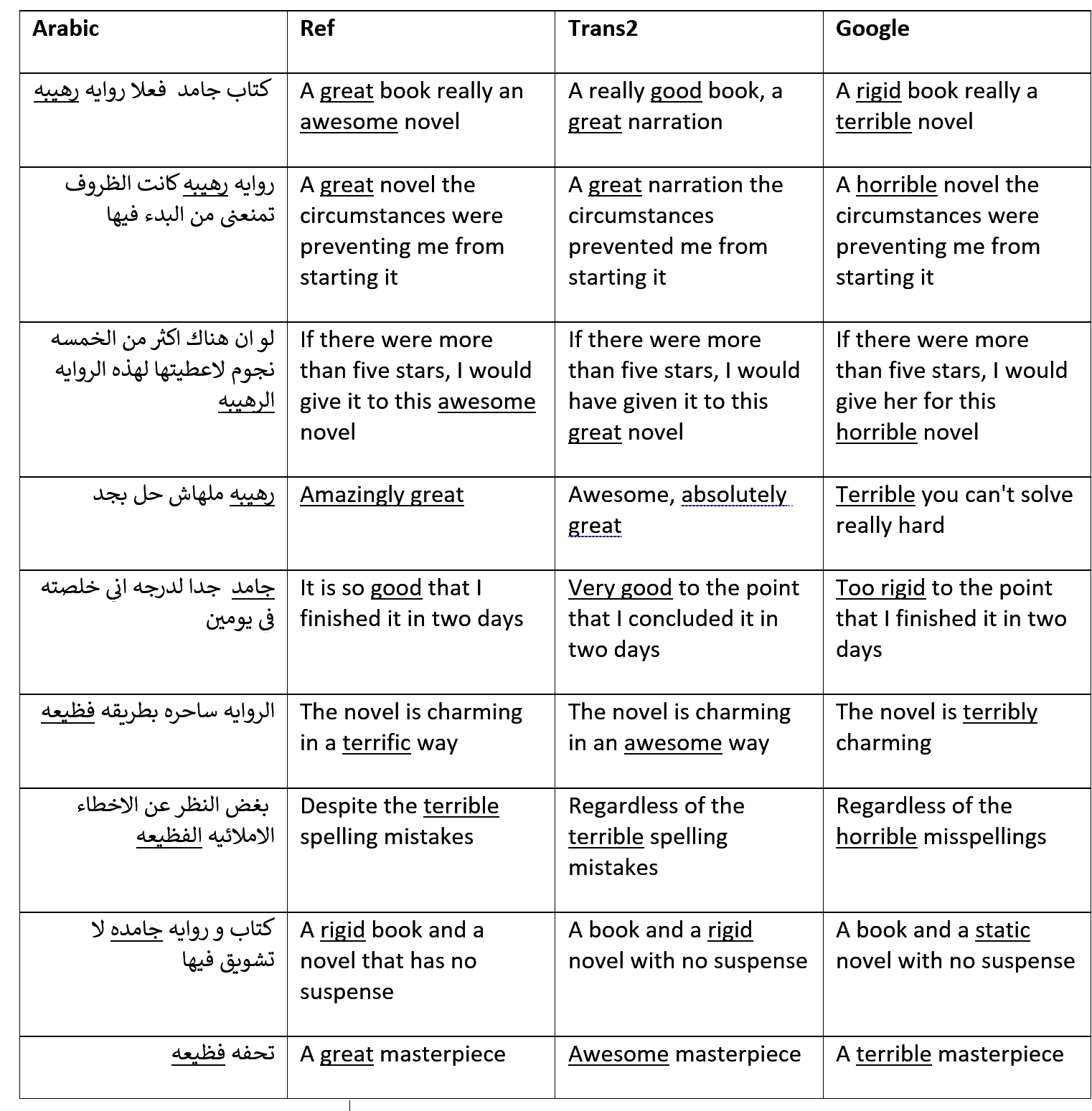}
\label{egtable}
\end{figure}

\newpage
\section{List of Most Frequent Contronyms}
\label{contr}
\begin{figure}[h]
\centering
\includegraphics[scale=.55,trim={0cm .05cm 0cm 0cm},clip]{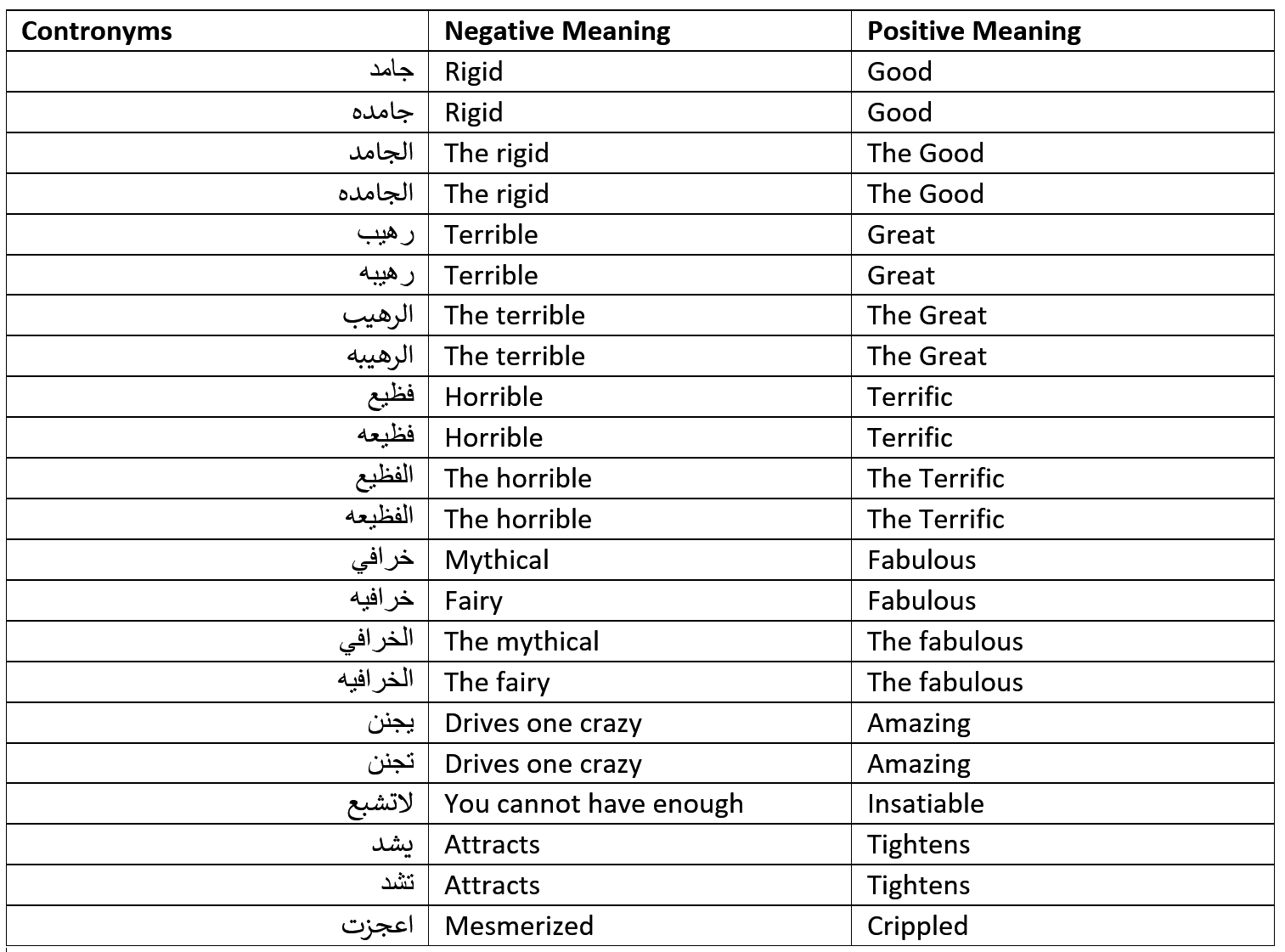}
\label{conttable}
\end{figure}

\end{appendices}

\end{document}